\documentclass[runningheads]{llncs}

\usepackage[T1]{fontenc}

\usepackage{graphicx}
\usepackage{amsmath}
\usepackage{latexsym}
\usepackage{amssymb}
\usepackage{amsfonts}

\usepackage{booktabs}
\usepackage{enumitem}
\usepackage{color}
\usepackage{algorithmic}
\usepackage[ruled]{algorithm2e}
\usepackage{subfigure}
\usepackage{makecell}

\RestyleAlgo{ruled}
\SetKwComment{Comment}{/* }{ */}

\usepackage{tabularray}
\usepackage{adjustbox}
\usepackage{multirow}
\usepackage{colortbl}
\usepackage{hhline}
\usepackage[normalem]{ulem}

\usepackage{bbding}

\begin{document}
\title{IFFair: Influence Function-driven Sample Reweighting for Fair Classification}
%

\author{Jingran Yang\inst{1}\orcidID{0009-0008-6406-9222} \and
Min Zhang\inst{2(}\Envelope\inst{)}\orcidID{0000-0002-3152-4347}
\and
Lingfeng Zhang\inst{1}\orcidID{0000-0002-6427-9587}
\and
Zhaohui Wang\inst{1}\orcidID{0009-0002-7774-5206}
\and
Yonggang Zhang\inst{3}}
\authorrunning{J. Yang et al.}
\institute{East China Normal University, Shanghai, 200062, P.R. China\\
\email{nancyyyyang@163.com}, \email{lanford217@gmail.com}, \email{sternstund22@gmail.com}\\
\and
Shanghai Key Laboratory of Trustworthy Computing, Shanghai, P.R. China\\
\email{mzhang@sei.ecnu.edu.cn}
\and
Key Laboratory of Symbolic Computation and Knowledge Engineering of Ministry of Education, Jilin University, Changchun, 120012, P.R. China\\
\email{zhangyg@jlu.edu.cn}
}
\maketitle              
\begin{abstract}
Because machine learning has significantly improved efficiency and convenience in the society, it's increasingly used to assist or replace human decision-making. However, the data-based pattern makes related algorithms learn and even exacerbate potential bias in samples, resulting in discriminatory decisions against certain unprivileged groups, depriving them of the rights to equal treatment, thus damaging the social well-being and hindering the development of related applications. Therefore, we propose a pre-processing method IFFair based on the influence function. Compared with other fairness optimization approaches, IFFair only uses the influence disparity of training samples on different groups as a guidance to dynamically adjust the sample weights during training without modifying the network structure, data features and decision boundaries. To evaluate the validity of IFFair, we conduct experiments on multiple real-world datasets and metrics. The experimental results show that our approach mitigates bias of multiple accepted metrics in the classification setting, including demographic parity, equalized odds, equality of opportunity and error rate parity without conflicts. It also demonstrates that IFFair achieves better trade-off between multiple utility and fairness metrics compared with previous pre-processing methods.
\end{abstract}
\section{Introduction}\label{sec:introduction}
With the rapid development of intelligent computing, machine learning has made tremendous progress in application from academia to industry. However, existing research suggests that it raises ethical issues that may replicate and exacerbate human bias in certain scenarios, such as finance \cite{khandani2010consumer,aseervatham2016unisex}, employment \cite{yarger2020algorithmic,dastin2022amazon,kiritchenko2018examining,guion1966employment,osoba2017intelligence}, medical diagnosis \cite{obermeyer2019dissecting,fang2020achieving,spanakis2013race}, recommendation \cite{datta2014automated,zhao2017men,ross2011women}, examinations \cite{cleary1966test,cleary1968test} and legal \cite{berk2021fairness}. These issues may further deteriorate negative impacts on unprivileged groups or individuals, thus trustworthy machine learning has been proposed as a core technology in artificial intelligence safety \cite{loi2019include,yapo2018ethical}.

As an important aspect of trustworthy learning, fairness refers to the neutrality of a system in decision-making and resource allocation, ensuring no bias based on inherent or acquired characteristics \cite{saxena2019fairness}. Such characteristics are called sensitive attributes, which are features related to people. Depending on the value of sensitive attributes, samples can be divided into different groups (e.g., male and female), which may be treated differently by algorithms.

Researches on group fairness are divided into pre-processing \cite{grgic2016case,feldman2015certifying,shorten2019survey,burnaev2015influence}, in-processing \cite{beutel2019putting,madras2018learning,zhao2019conditional} and post-processing \cite{kamiran2009classifying,fish2016confidence}. In-processing and post-processing mechanisms mitigate bias by adjusting models or outputs. However, the emergence of biased algorithms usually because they learn from data, so pre-processing methods are proposed. Some strategies like resampling \cite{cui2019class,burnaev2015influence}, augmentation \cite{shorten2019survey} and data synthesis \cite{chawla2002smote} cause data modification, which is unacceptable in some cases \cite{barocas2016big}. Therefore, we propose IFFair (\underline{I}nfluence \underline{F}unction-driven Sample Reweighting for \underline{Fair} Classification) without modifying data, network or output, just reweights data via influence function \cite{ling1984residuals}. Our contributions include:
\begin{itemize}
    \item IFFair has the pre-processing advantage without modifing model structures and outputs. Based on the Hessian matrix and loss gradient, it can quantify the group-oriented influence without disturbing original data distribution.
    \item Compared with vanilla classifiers, the prediction after reweighting via IFFair has better performance on 7 datasets and 4 fairness metrics without conflict.
    \item Our method ensures a better trade-off between 4 fairness and 3 utility metrics than previous pre-processing work while improving group fairness.
    \item IFFair is a general method which is not only suitable for LR used in previous work, but also performs well on DNN. It's a property that some comparable pre-processing methods cannot guarantee.
\end{itemize}
\section{Preliminary}\label{sec:preliminary}
In this section, we will formally define and explain the notations, definitions and introduce the fair classification problem settings explored in this paper.
\subsection{Notation}\label{subsec:notion}
Consider a general task to build a model $f$ from the input space $\mathcal{X}$ to output space $\mathcal{Y}$ based dataset $\mathcal{Z} = \{(x_i,y_i)\}_{i=1}^n$, uppercase letters $X$, $S$, $A$ and $Y$ used to represent the random variables, where $A$ is partial of $X$ called sensitive attribute. To simplify the representation, we assume $A,Y \in \{0,1\}$. On this basis, the predicted label based on corresponding probability of $X$ is $\hat{Y}=\operatorname*{arg\,max}f(X)$. Assume $\theta$ is parameters of model $f$, thus the loss function of $z$ is $\mathcal{L}(z,\theta)$, the average loss on the training set $\mathcal{Z}$ is empirical risk $R(\theta)$. Thus the optimal model parameter $\hat{\theta}$ can be obtained by Empirical Risk Minimization (ERM).

\subsection{Evaluation Metrics}\label{subsec:metrics}
The core perspectives of model evaluation include fairness and utility, thus we give some accepted definitions and metrics here.
\subsubsection{Utility Metrics}\label{subsubsec:util_metrics}
Utility reflects the output quality of classifiers and usually involve ollowing key metrics, where higher value mean the better performance.
\begin{definition}[Accuracy]\label{def:acc}
Accuracy is the ratio of the counts of data correctly predicted by the model to all, which is calculated as $Acc = P(\hat{Y}=Y)$.
\end{definition}

\begin{definition}[F1-score]\label{def:f1}
F1-score measures a harmonic mean of precision and recall, which is calculated as $F1=2\times\frac{Precision\times Recall}{Precision+Recall}$.
\end{definition}

\begin{definition}[ROC Curve]\label{def:roc}
ROC (Receiver Operating Characteristic) Curve shows the performance of the model under different thresholds by plotting the relationship between TPR and FPR in the binary classification problem.
\end{definition}

\begin{definition}[AUC]\label{def:auc}
AUC (Area Under the ROC Curve) measures the area between the space under the ROC Curve and the coordinate axis.
\end{definition}

\subsubsection{Fairness Metrics}\label{subsubsec:fair_metrics}
Based on various scenarios, different fairness definitions and metrics have been proposed. We focus on the group fairness which is widely studied, aims to eliminate bias across subgroups just divided by sensitive attributes. 

\begin{definition}[Demographic Parity, DP]\label{def:dp}
DP enforces the prediction and sensitive attribute are statistically independent, i.e. $P(\hat{Y}=1|A=1) = P(\hat{Y}=1|A=0)$, whose metric is $\Delta DP = P(\hat{Y}=1|A=1) - P(\hat{Y}=1|A=0)$.
\end{definition}

\begin{definition}[Equality of Opportunity, EOP]\label{def:fpr}
EOP enforces the equality on the false positive rate across different groups, whose metric is $\Delta FPR = P(\hat{Y}=1|A=1,Y=0) - P(\hat{Y}=1|A=0,Y=0)$.
\end{definition}

\begin{definition}[Equalized Odds, EOdds]\label{def:eodds}
EOdds requires both FPR and TPR of predictions between unprivileged and privileged groups are same, it can be evaluate by $\Delta EOdds=\frac{1}{2}(|P(\hat{Y}=1|A=1,Y=0)-P(\hat{Y}=1|A=0,Y=0)|+|P(\hat{Y}=1|A=1,Y=1)-P(\hat{Y}=1|A=0,Y=1)|)$.
\end{definition}

\begin{definition}[Error Gap]\label{def:err_gap}
Given a data distribution $\mathcal{Z}$, the Error Gap of a classifier is $\Delta Err = | P(\hat{Y}\neq Y|A=1) - P(\hat{Y}\neq Y|A=0)|.$
\end{definition}

Differ from above classic metrics, Error Gap is defined based on the perspective of performance disparity between groups \cite{zhao2019conditional}. However, for all fairness metrics, the larger absolute value means more severe bias or discrimination exists, and the complete fairness is satisfied when the value is equal to 0.

\subsection{Formalization of Influence Function}\label{subsec:IF}
The influence function is a statistical learning concept in robustness research, which evaluates the changing rate in model parameter estimation when a specific sample weights are slightly perturbed. It is calculated by gradient and Hessian matrix of the loss function \cite{koh2017understanding}.

Consider a classification model $f(\theta)$ with the parameter vector $\theta$, which is trained on a training set $\mathcal{Z} = \{z_i\}_{i=1}^n$. Given a loss function $\mathcal{L}(z,\theta)$, then the empirical risk is $R(\theta) = \frac{1}{n} \sum_{i=1}^{n}\mathcal{L}(z_i, \theta)$. Thus the optimal model parameters is
\begin{equation}\label{eq:theta_erm}
    \hat{\theta} = \arg\min_{\theta} R(\theta) = \arg\min_{\theta} \frac{1}{n} \sum_{i=1}^{n}\mathcal{L}(z_i, \theta).
\end{equation}
For a weight increment $\epsilon$ used to perturb the weight of $z$, the new empirical risk function becomes $R(\theta) + \epsilon \mathcal{L}(z, \theta)$, and the new optimal parameter is
\begin{equation}\label{eq:theta_ez}
\begin{aligned}
    \hat{\theta}_{\epsilon,z} =\arg\min_{\theta} ( \frac{1}{n} \sum_{i=1}^{n}\mathcal{L}(z_i, \theta) + \epsilon \mathcal{L}(z, \theta)).
\end{aligned}
\end{equation}
Suppose $\Delta_{\epsilon} = \hat{\theta}_{\epsilon,z} - \hat{\theta}$ is used to measure the change of $\theta$, because $\hat{\theta}$ is independent of $\epsilon$, we have $\frac{d\hat{\theta}_{\epsilon,z}}{d\epsilon} = \frac{d\Delta_\epsilon}{d\epsilon}$. Therefore, the influence function is
\begin{equation}\label{eq:IF}
    \mathcal{I}_{\mathcal{L}}(z) = \frac{d\hat{\theta}_{\epsilon,z}}{d\epsilon}\bigg|_{\epsilon=0}=\frac{d\Delta\epsilon}{d\epsilon}\bigg|_{\epsilon=0} = -H_{\hat{\theta}}^{-1} \nabla_\theta \mathcal{L}(z, \hat{\theta}),
\end{equation}
which evulates the relationship between changes in model parameters and changes in the weights of sample. $H_{\hat{\theta}}$ is Hessian matrix measuring the average loss of a dataset, which is calculated as the second-order partial derivative of $R(\theta)$ at $\hat{\theta}$:
\begin{equation}\label{eq:hessian}
    H_{\hat{\theta}} = \nabla_{\theta}^2 R(\hat{\theta}) = \frac{1}{n} \sum_{i=1}^{n} \nabla_{\theta}^2 \mathcal{L}(z_i, \hat{\theta}),
\end{equation}

\section{IFFair: Influence Function-driven Sample Reweighting}\label{sec:IFFair}
Group fairness aims to ensure subgroups divided by sensitive attributes receive equal treatment. However, it's shown that biases often originates from data.
Thus, we propose IFFair to quantify a sample influence on subgroups and reweights them to explain and mitigate biases in data.

\subsection{Quantify Group-oriented Influence}\label{subsec:IF_group}
Consider a binary classification task with target variable $Y$ and sensitive attribute $A\in \{0,1\}$, dataset $\mathcal{Z}$ is divided into privileged group $\mathcal{Z}_1 = \{z_i \mid A_i = 1\}$ and unprivileged group $\mathcal{Z}_0 = \{z_i \mid A_i = 0\}$. Group unfairness indicates that the decision is susceptible to $A$, resulting a negative decision $\hat{Y}^-$ to $\mathcal{Z}_0$ and a positive decision $\hat{Y}^+$ to $\mathcal{Z}_1$. Following the definition of influence function in Equation \eqref{eq:IF}, which measures the impact on the model parameters after removing $z$, and the model parameters change is $\Delta\theta = H_{\hat{\theta}}^{-1}\nabla_{\theta}\mathcal{L}(z, \hat{\theta})$. Therefore, the loss change reflecting the influence of $z$ on $z_j$ affected by $\Delta\theta$ is
\begin{equation}\label{eq:delta_lj}
    \Delta\mathcal{L}_j=\nabla_{\theta}\mathcal{L}(z_j,\hat{\theta})^T \cdot \Delta\theta
    =\nabla_{\theta}\mathcal{L}(z_j,\hat{\theta})^T \cdot H_{\hat{\theta}}^{-1}\nabla_{\theta}\mathcal{L}(z,\hat{\theta}).
\end{equation}

For IFFair that focuses on group fairness, so we extend the individual-oriented influence $\mathcal{I}(z)$ to group-oriented influence $\mathcal{I}_{a}(z)$, which aggregates the influence of $z$ on all samples within the subgroup $\mathcal{Z}_a$:
\begin{equation}\label{eq:IF_group}
    \mathcal{I}_a(z) = \sum_{z_j \in \mathcal{Z}_a} \nabla_{\theta}\mathcal{L}(z_j,\hat{\theta})^T \cdot H_{\hat{\theta}}^{-1}\nabla_{\theta}\mathcal{L}(z,\hat{\theta}).
\end{equation}
The details of calculation is shown in Algorithm \ref{alg:IF_group}.

\subsection{IFFair with Trade-off Constraints}
\subsubsection{Variants of IFFair}
To improve group fairness via data reweighting, we propose two variants of IFFair distinguished by reweighting strategies. Both variants rely on group-oriented influence from Algorithm \ref{alg:IF_group} to identify biased samples in $\mathcal{Z}_{bias}$.

IFFair-Uniform enforces $\mathcal{W}_{bias}=\{w_i|z_i \in \mathcal{Z}_{bias}\}$ is equal to an uniform new weight $w'$, where $w'\in [0, 1]$. It assumes that samples in $\mathcal{Z}_{bias}$ have the consistent influence on fairness, simplifies the optimization process by reducing the number of decision variables $w'$. However, its uniformity assumption may overlook the nuanced differences in individual sample influences. For example, different samples in $\mathcal{Z}_{bias}$ might drive unfairness to different degrees, but IFFair-Uniform treats them equally, potentially limiting the maximum achievable fairness.  
IFFair-Diverse proposed to solve this concern, it assigns adaptive weights $w_i'=1-\Delta w_i$ to each biased sample $z_i \in \mathcal{Z}_{bias}$, enabling precise adjustment of how each sample influence to fairness. This fine-grained weighting strategy captures more detailed differences in discrimination, allowing samples with varying degrees of discrimination to receive different levels of attention, thereby achieving better fairness performance. However, the increasing number of decision variables $\Delta w_i$ may lead more complexity for linear programming.

\subsubsection{Trade-off Constraints between Fairness and Utility}
There is a widespread belief in the research community that there is a inevitable trade-off between fairness and utility \cite{wick2019unlocking,haas2019price,berk2021fairness}. Therefore, when evaluating bias mitigation methods, we must not only consider their effectiveness in optimizing fairness but also take into account the decision performance. To balance fairness and utility in IFFair, we incorporate constraints during the process of data reweighting. These constraints ensure that while we aim to minimize fairness disparities, the degradation of model utility is kept within an acceptable range.
\begin{algorithm}[h]
\caption{Calculation of group-oriented influence}\label{alg:IF_group}
\KwIn{Dataset $\mathcal{Z}=\{z_i\}_{i=1}^n$, Trained model $f$ with optimal parameter $\hat{\theta}$}
\KwOut{Group-oriented influence of $z_i \in \mathcal{Z}$}

\Comment{Partition data by sensitive attribute}
$\mathcal{Z}_a = \{z_i \mid A_i = a\}_{i=1}^{n_a}, \quad a\in \{0,1\}$\\

\Comment{Calculate individual-oriented gradient and group-oriented cumulative gradient}
$\nabla_{\theta}\mathcal{L}(z_i,\hat{\theta}) = (f(x_i,\hat{\theta}) - y_i) \cdot \nabla_{\theta}f(x_i,\hat{\theta}) \quad \forall z_i \in \mathcal{Z}$\\
$\nabla R_a = \sum_{z_i \in \mathcal{Z}_a} \nabla_{\theta}\mathcal{L}(z_i,\hat{\theta}),\quad a\in\{0,1\}$\\

\Comment{Compute inverse Hessian-vector products (IHVP) for subgroups}
$H_{\hat{\theta}} = \nabla^2_{\theta} \frac{1}{n}\sum_{z_i \in \mathcal{Z}} \mathcal{L}(z_i,\hat{\theta})$\\
$IHVP_0 = H_{\hat{\theta}}^{-1} \nabla R_0,\quad IHVP_1 = H_{\hat{\theta}}^{-1} \nabla R_1$

\Comment{Calculate group-oriented influence}
\For{$z_i \in \mathcal{Z}$}{
    $\mathcal{I}_0(z_i) = \nabla_{\theta}\mathcal{L}(z_i,\hat{\theta}) \cdot IHVP_0$\\
    $\mathcal{I}_1(z_i) = \nabla_{\theta}\mathcal{L}(z_i,\hat{\theta}) \cdot IHVP_1$\\
}
\algorithmicreturn{ $\mathcal{I}_0(z_i),\mathcal{I}_1(z_i)  \quad \forall z_i \in \mathcal{Z}$ }
\end{algorithm}

For IFFair-Uniform, we optimize a scalar $w'$ applied uniformly to $\mathcal{Z}_{bias}$, balancing fairness and utility via constraint \eqref{eq:IFFair-Uniform_constraint}. We first normalize different fairness metrics to get $S_{fair} = \sum_{m=1}^{M} \frac{|fair_m|}{|base_m| + \epsilon}$ evaluated the fairness disparity, where $M$ is fairness metrics count, $fair_m$ is the actual value of $m$-th fairness metric, $base_m$ is basic value of the vanilla model. The goal of the constrainted-optimization in IFFair-Uniform is to minimize fairness score $S_{fair}$ while preserving utility:  
\begin{equation}\label{eq:IFFair-Uniform_constraint}
    \begin{array}{ll}
        \text{minimize} & S_{fair}(w'), \quad w' \in [0, 1]\\
        \text{subject to} & util(w') \geq util_{base} \cdot (1 - \tau).
    \end{array}
\end{equation}
$util(w')$ denotes new values of utility metrics of the reweighted model, $\tau$ limits the utility degradation in an acceptable range, and the piecewise linear (PWL) approximation is used to get the optimal parameter $w^*$ of $w'$.

For IFFair-Diverse, we optimize the sample-specific adjustment weight $\Delta w_i \in [0,1]$ where $\Delta IF(z_i) < 0$, which indicates $z_i$ has a negative impact on model fairness. The goal is to obtain the optimal weight $w_i = 1 - \Delta w_i$, where $\Delta w_i = 1$ means completely removing $z_i$. Specifically, we first calculate the group fairness optimization potential and utility optimization potential as $max\_fair= \sum_{\Delta IF(z_i)<0}\Delta IF(z_i)$ and $max\_util=\sum_{IF(z_i)<0}IF(z_i)$. Then the constrained-optimization process of IFFair-Diverse is:  
\begin{equation}\label{eq:IFFair-Diverse_constraint}
    \begin{array}{ll}
        \text{minimize} & \sum_{i=1}^n \Delta w_i, \quad \Delta w_i \in [0, 1]\\
        \text{subject to} & \sum_{i=1}^n \Delta IF(z_i) \cdot \Delta w_i \leq \lambda_f \cdot max\_fair, \\
        & \sum_{i=1}^n IF(z_i) \cdot \Delta w_i \leq \lambda_u \cdot max\_util.
    \end{array}
\end{equation}

It's committed to maintaining the original data distribution while optimizing group fairness, and making better trade-off between fairness and utility. In Equaltion \eqref{eq:IFFair-Diverse_constraint}, $\lambda_f,\lambda_u \in [0,1]$ control the max fairness improvement and utility performance. Finally, we derive the final weights $w_i^* = 1 - \Delta w_i^*$ where $\Delta w_i^*$ is the optimal adjustment factor for $z_i \in \mathcal{Z}_{diverse}$. Details of IFFair based on the group-oriented influence and trade-off constraints are shown in Algorithm \ref{alg:IF_reweight}.
\begin{algorithm}[h]
\caption{IFFair: Influence Function-driven Sample Reweighting}\label{alg:IF_reweight}
\KwIn{Equal weighted dataset $\mathcal{Z}$, Vanilla model $f$, Utility threshold $\tau$}
\KwOut{Optimized weights set $\mathcal{W}^*$, Fair model $f^*$}

\Comment{Calculate group-oriented influence of subgroups with different sensitive attributes}
Compute $\mathcal{I}_0(z_i), \mathcal{I}_1(z_i)$ for all $z_i \in \mathcal{Z}$ using Algorithm \ref{alg:IF_group}\\
\Comment{Identify the biased samples}
Identify biased samples in IFFair-Uniform variant $\mathcal{Z}_{uniform} =\{z_i \in \mathcal{Z}|(\mathcal{I}_0(z_i) < 0) \land (\mathcal{I}_1(z_i) > 0)\} $\\
Calculate group-oriented influence disparity $\Delta IF(z_i)=\mathcal{I}_0(z_i)-\mathcal{I}_1(z_i), \forall z_i \in \mathcal{Z}$\\
Identify biased samples in IFFair-Diverse variant $\mathcal{Z}_{diverse}=\{z_i \in \mathcal{Z}|\Delta IF(z_i)<0\}$

\Comment{Solve constrained weight optimization of two IFFair variants}
\If{IFFair-Uniform}{
    Define a decision variable $w' \in [0, 1]$ that is going to be uniformly reassigned to $\forall z_i \in \mathcal{Z}_{uniform}$\\
    Minimize optimization objective with constraint as Equation \eqref{eq:IFFair-Uniform_constraint} to obtain the optimal weight $w^*$, which is under the condition of accepted utility degradation $\tau$\\
    Update vanilla weights to $\mathcal{W}^* = \{ w^* \mid \forall z_i \in \mathcal{Z}_{uniform}, w_i^* = w^*; \forall z_i \notin \mathcal{Z}_{uniform}, w_i^* = 1 \}$
}
\ElseIf{IFFair-Diverse}{
    Define adaptive decision variables $\Delta w_i \in [0, 1]$ used to update vanilla weights $w_i$ for $\forall z_i \in \mathcal{Z}_{diverse}$\\
    Minimize optimization objective with constraint as Equation \eqref{eq:IFFair-Diverse_constraint} to obtain optimal adjusted weights $\Delta w_i^*$\\
    Update vanilla weights to $\mathcal{W}^* = \{ w^* \mid \forall z_i \in \mathcal{Z}_{diverse}, w_i^* =1-\Delta w_i^*; \forall z_i \notin \mathcal{Z}_{diverse}, w_i^* = 1 \}$
}

\Comment{Obtain the fair model with optimized weights}
Retrain the model $f^*$ on $\mathcal{Z}$ with the optimized weights set $\mathcal{W}^*$\\

\algorithmicreturn{ $\mathcal{W}^*$, $f^*$ }
\end{algorithm}
\section{Experiment}
\subsection{Experimental Settings}
\subsubsection{Benchmark Datasets}
We perform experiments on 7 popular real-world datasets usually used in the previous research of ML fairness as Table \ref{tab:7datasets}.
\subsubsection{Baseline Models}
IFFair is a pre-processing method applying fairness constraints on training set to mitigate bias before constructing models. To verify its effectiveness, we select 7 pre-processing methods. \textbf{Suppression} \cite{grgic2016case} removes the feature column where sensitive attribute locating. \textbf{IPW} \cite{hofler2005use} adjusts weights that are inversely proportional to the probability of their groups divided on the sensitive attribute or its joint distribuution with label. \textbf{DiscriminationFree} \cite{calders2010three} includes 3 types: ModifiedNB modifies the probability distribution. 2NB trains and balanced two independent models for different values of sensitive attribute. LatentV optimizes a latent unbiased label by expectation maximization. \textbf{FairMap} \cite{calmon2017optimized} achieves unbiased data transformation by learning the stochastic mapping function $P_{\hat{X},\hat{Y}|X,Y,A}$. \textbf{CostFree} \cite{li2022achieving} models the influence of dataset to the classifier based on specific fairness and utility losses, and designs dataset-oriented coefficients to trade off them. \textbf{LabelBias} \cite{jiang2020identifying} iteratively learns sample weights based on constraints ensuring data more heavily affected by bias receive more adjustments. \textbf{ARL} \cite{lahoti2020fairness} identifies regions with high loss and assigns higher weights to them by max-min optimization.
\begin{table}[!]
\centering
\caption{Statistics of benchmark datasets.}
\label{tab:7datasets}
\resizebox{0.95\textwidth}{!}{
\begin{tabular}{|c|c|c|c|c|c|c|c|} 
\hline
Dataset name              & German      & Bank       & Adult     & LSAC & MEPS     & COMPAS        & Comm           \\ 
\hline
Sensitive attribute & Age         & Age        & Sex       & Sex  & Race     & Race          & Race           \\
\hline
Favoriable label    & Good credit & Subscriber & Income>50k & Pass & Utilizer & No recidivism & Lower violent  \\
\hline
\end{tabular}
}
\end{table}

To ensure the comparability of experimental results, all baseline networks are attempted to be observed in the same experimental settings as IFFair, that is, on the same evulation metrics, basic networks and benchmark datasets.

\subsection{Experimental Results and Analysis}
\subsubsection{RQ1: How well does IFFair mitigate bias of the vanilla model?}
As a pre-processing method, it's necessary to eliminate the data bias. Specifically, we analyze the evaluation results from two aspects. \textbf{RQ1.1:} Whether IFFair improve fairness compared with the vanilla model? \textbf{RQ1.2:} Different fairness may contradict each other \cite{berk2021fairness,du2020fairness}, so we discuss whether there is an unexpected conflict after using IFFair. The main classifier of RQ1 is LR as other pre-processing methods, and experimental results are recorded in Table \ref{tab:fair_U&D_LR}.
\begin{table}[!]
\centering
\caption{Fairness performance of IFFair methods compared with baselines on LR. For methods in cells, bold texts represent the best fairness, black cells indicate deterioration for $Original$, light gray cells indicate IFFair is inferior to them, '\textbackslash' indicates invalidity.}
\label{tab:fair_U&D_LR}
\begin{adjustbox}{rotate=90, max height=\textheight}
\begin{tabular}{|c|c|ccccccc|c|c|ccccccc|} 
\hline
\multirow{2}{*}{Fairness}        & \multirow{2}{*}{Methods} & \multicolumn{7}{c|}{Datasets}                                                                                                                                                                                                                                                                                                                                                       & \multicolumn{1}{c}{\multirow{2}{*}{Fairness}} & \multicolumn{1}{c}{\multirow{2}{*}{Methods}} & \multicolumn{7}{c|}{Datasets}                                                                                                                                                                                                                                                                                                                                                \\ 
\cline{3-9}\cline{12-18}
                                 &                          & Bank                                          & German                                              & LSAC                                                & Adult                                               & COMPAS                                               & Comm                                                & MEPS                                                 & \multicolumn{1}{c}{}                          & \multicolumn{1}{c}{}                         & Bank                                         & German                                              & LSAC                                          & Adult                                               & COMPAS                                              & Comm                                                 & MEPS                                                 \\ 
\hline
\multirow{16}{*}{$\Delta DP$}    & Original                 & -0.0756                                       & 0.1719                                              & 0.0457                                              & 0.2131                                              & 0.1519                                               & 0.4265                                              & 0.1906                                               & \multirow{16}{*}{$\Delta FPR$}                & Original                                     & 0.0000                                       & 0.1839                                              & 0.0264                                        & 0.0881                                              & 0.0637                                              & 0.1454                                               & 0.0739                                               \\
                                 & Supreesion               & -0.0700                                       & 0.1467                                              & {\cellcolor[rgb]{0.749,0.749,0.749}}0.0152          & 0.2013                                              & {\cellcolor{black}}\textcolor{white}{0.1531}         & 0.4159                                              & 0.1679                                               &                                               & Supreesion                                   & 0.0000                                       & 0.1722                                              & -0.0041                                       & 0.0697                                              & 0.0637                                              & 0.1254                                               & 0.0451                                               \\
                                 & IPW-S                    & {\cellcolor{black}}\textcolor{white}{0.5511}  & 0.0973                                              & {\cellcolor{black}}\textcolor{white}{-0.1904}       & {\cellcolor{black}}\textcolor{white}{0.2634}        & {\cellcolor{black}}\textcolor{white}{0.1857}         & 0.4110                                              & {\cellcolor[rgb]{0.749,0.749,0.749}}-0.0274          &                                               & IPW-S                                        & {\cellcolor{black}}\textcolor{white}{0.3400} & 0.1345                                              & {\cellcolor{black}}\textcolor{white}{-0.2136} & {\cellcolor{black}}\textcolor{white}{0.1676}        & {\cellcolor{black}}\textcolor{white}{0.0941}        & 0.1254                                               & {\cellcolor{black}}\textcolor{white}{-0.1259}        \\
                                 & IPW-SY                   & {\cellcolor{black}}\textcolor{white}{0.3261}  & 0.1304                                              & -0.0461                                             & {\cellcolor{black}}\textcolor{white}{0.2473}        & {\cellcolor[rgb]{0.749,0.749,0.749}}0.0083           & {\cellcolor[rgb]{0.749,0.749,0.749}}0.2963          & {\cellcolor[rgb]{0.749,0.749,0.749}}\textbf{-0.0172} &                                               & IPW-SY                                       & {\cellcolor{black}}\textcolor{white}{0.2500} & 0.1650                                              & {\cellcolor{black}}\textcolor{white}{-0.0651} & 0.0195                                              & -0.0510                                             & {\cellcolor[rgb]{0.749,0.749,0.749}}\textbf{-0.0022} & {\cellcolor{black}}\textcolor{white}{-0.1057}        \\
                                 & ModifiedNB               & -0.0756                                       & 0.1018                                              & {\cellcolor[rgb]{0.749,0.749,0.749}}0.0160          & {\cellcolor{black}}\textcolor{white}{0.6003}        & 0.1464                                               & {\cellcolor[rgb]{0.749,0.749,0.749}}0.3381          & {\cellcolor{black}}\textcolor{white}{0.2178}         &                                               & ModifiedNB                                   & 0.0000                                       & 0.1316                                              & 0.0021                                        & {\cellcolor{black}}\textcolor{white}{0.5594}        & {\cellcolor{black}}\textcolor{white}{0.0673}        & 0.0773                                               & {\cellcolor[rgb]{0.749,0.749,0.749}}0.0326           \\
                                 & 2NB                      & 0.0047                                        & 0.0629                                              & 0.0299                                              & 0.1875                                              & {\cellcolor{black}}\textcolor{white}{0.2121}         & {\cellcolor{black}}\textcolor{white}{0.4419}        & {\cellcolor[rgb]{0.749,0.749,0.749}}0.0964           &                                               & 2NB                                          & {\cellcolor{black}}\textcolor{white}{0.0300} & 0.0621                                              & 0.0143                                        & {\cellcolor{black}}\textcolor{white}{0.1520}        & {\cellcolor{black}}\textcolor{white}{0.1345}        & 0.1451                                               & {\cellcolor{black}}\textcolor{white}{0.1286}         \\
                                 & LatentV                  & {\cellcolor{black}}\textcolor{white}{-0.0874} & 0.1062                                              & {\cellcolor[rgb]{0.749,0.749,0.749}}0.0155          & {\cellcolor[rgb]{0.749,0.749,0.749}}0.1473          & 0.0981                                               & 0.3655                                              & {\cellcolor[rgb]{0.749,0.749,0.749}}0.0928           &                                               & LatentV                                      & 0.0000                                       & 0.1549                                              & 0.0042                                        & -0.0262                                             & 0.0148                                              & 0.0786                                               & {\cellcolor[rgb]{0.749,0.749,0.749}}0.0315           \\
                                 & FairMap                  & $\textbackslash$                              & $\textbackslash$                                    & $\textbackslash$                                    & {\cellcolor[rgb]{0.749,0.749,0.749}}\textbf{0.0024} & {\cellcolor[rgb]{0.749,0.749,0.749}}\textbf{-0.0042} & $\textbackslash$                                    & $\textbackslash$                                     &                                               & FairMap                                      & $\textbackslash$                             & $\textbackslash$                                    & $\textbackslash$                              & 0.0102                                              & -0.0374                                             & $\textbackslash$                                     & $\textbackslash$                                     \\
                                 & CostFree-DP              & $\textbackslash$                              & {\cellcolor[rgb]{0.749,0.749,0.749}}\textbf{0.0054} & $\textbackslash$                                    & {\cellcolor[rgb]{0.749,0.749,0.749}}-0.0504         & 0.1188                                               & {\cellcolor[rgb]{0.749,0.749,0.749}}0.3370          & $\textbackslash$                                     &                                               & CostFree-DP                                  & $\textbackslash$                             & 0.0459                                              & $\textbackslash$                              & {\cellcolor{black}}\textcolor{white}{-0.3717}       & 0.0326                                              & 0.1181                                               & $\textbackslash$                                     \\
                                 & CostFree-FPR             & $\textbackslash$                              & 0.1719                                              & $\textbackslash$                                    & 0.1656                                              & 0.1433                                               & 0.4110                                              & $\textbackslash$                                     &                                               & CostFree-FPR                                 & $\textbackslash$                             & 0.1839                                              & $\textbackslash$                              & -0.0027                                             & 0.0407                                              & {\cellcolor{black}}\textcolor{white}{0.1522}         & $\textbackslash$                                     \\
                                 & LabelBias-DP             & -0.0756                                       & -0.0154                                             & {\cellcolor[rgb]{0.749,0.749,0.749}}-0.0171         & -0.1598                                             & -0.1743                                              & {\cellcolor[rgb]{0.749,0.749,0.749}}\textbf{0.1731} & {\cellcolor{black}}\textcolor{white}{-0.5527}        &                                               & LabelBias-DP                                 & 0.0000                                       & {\cellcolor[rgb]{0.749,0.749,0.749}}\textbf{0.0081} & {\cellcolor{black}}\textcolor{white}{-0.0369} & {\cellcolor{black}}\textcolor{white}{-0.4406}       & {\cellcolor{black}}\textcolor{white}{-0.2396}       & -0.0614                                              & {\cellcolor{black}}\textcolor{white}{-0.5208}        \\
                                 & LabelBias-FPR            & -0.0631                                       & 0.1097                                              & {\cellcolor[rgb]{0.749,0.749,0.749}}\textbf{0.0119} & 0.1790                                              & {\cellcolor[rgb]{0.749,0.749,0.749}}0.0180           & {\cellcolor[rgb]{0.749,0.749,0.749}}0.3228          & {\cellcolor[rgb]{0.749,0.749,0.749}}0.0753           &                                               & LabelBias-FPR                                & 0.0000                                       & 0.1417                                              & -0.0070                                       & 0.0095                                              & -0.0634                                             & {\cellcolor[rgb]{0.749,0.749,0.749}}0.0186           & -0.0700                                              \\
                                 & LabelBias-EOdds          & 0.0476                                        & -0.1023                                             & {\cellcolor{black}}\textcolor{white}{0.9291}        & {\cellcolor[rgb]{0.749,0.749,0.749}}0.1463          & {\cellcolor[rgb]{0.749,0.749,0.749}}0.0180           & 0.4198                                              & 0.1542                                               &                                               & LabelBias-EOdds                              & 0.0000                                       & -0.0689                                             & {\cellcolor{black}}\textcolor{white}{0.9471}  & -0.0815                                             & -0.0634                                             & 0.1376                                               & {\cellcolor[rgb]{0.749,0.749,0.749}}\textbf{0.0112}  \\
                                 & ARL                      & -0.0703                                       & 0.1383                                              & {\cellcolor[rgb]{0.749,0.749,0.749}}0.0166          & 0.1991                                              & 0.1512                                               & 0.4110                                              & 0.1691                                               &                                               & ARL                                          & 0.0000                                       & 0.1490                                              & -0.0030                                       & 0.0667                                              & 0.0546                                              & 0.1254                                               & 0.0556                                               \\ 
\cline{2-9}\cline{11-18}
                                 & IFFair-Uniform           & \textbf{-0.0024}                              & 0.0183                                              & 0.0222                                              & 0.1657                                              & 0.0455                                               & 0.3763                                              & 0.1906                                               &                                               & IFFair-Uniform                               & 0.0000                                       & 0.0314                                              & 0.0037                                        & 0.0022                                              & \textbf{0.0009}                                     & 0.0649                                               & 0.0739                                               \\
                                 & IFFair-Diverse           & 0.0039                                        & -0.0109                                             & 0.0197                                              & 0.1520                                              & 0.0709                                               & 0.3571                                              & 0.1273                                               &                                               & IFFair-Diverse                               & 0.0000                                       & -0.0108                                             & \textbf{0.0002}                               & \textbf{-0.0003}                                    & 0.0105                                              & 0.0581                                               & 0.0339                                               \\ 
\hline
\multirow{14}{*}{$\Delta EOdds$} & Original                 & 0.0816                                        & 0.2477                                              & 0.1119                                              & 0.1984                                              & 0.2465                                               & 0.3227                                              & 0.2150                                               & \multirow{14}{*}{$\Delta Err$}                & Original                                     & 0.0809                                       & 0.0925                                              & 0.0111                                        & 0.1102                                              & 0.0253                                              & 0.0249                                               & 0.1065                                               \\
                                 & Supreesion               & 0.0759                                        & 0.1755                                              & 0.0565                                              & 0.1697                                              & {\cellcolor{black}}\textcolor{white}{0.2489}         & 0.3063                                              & 0.1634                                               &                                               & Supreesion                                   & 0.0754                                       & {\cellcolor{black}}\textcolor{white}{0.1009}        & {\cellcolor{black}}\textcolor{white}{0.0131}  & 0.1056                                              & {\cellcolor{black}}\textcolor{white}{0.0265}        & {\cellcolor{black}}\textcolor{white}{0.0251}         & 0.0940                                               \\
                                 & IPW-S                    & {\cellcolor{black}}\textcolor{white}{0.8956}  & 0.2027                                              & 0.3322                                              & {\cellcolor{black}}\textcolor{white}{0.3227}        & {\cellcolor{black}}\textcolor{white}{0.3138}         & 0.2888                                              & 0.2090                                               &                                               & IPW-S                                        & {\cellcolor{black}}\textcolor{white}{0.5287} & {\cellcolor{black}}\textcolor{white}{0.2009}        & {\cellcolor{black}}\textcolor{white}{0.1854}  & {\cellcolor{black}}\textcolor{white}{0.1275}        & 0.0189                                              & {\cellcolor{black}}\textcolor{white}{0.0300}         & {\cellcolor[rgb]{0.749,0.749,0.749}}\textbf{0.0364}  \\
                                 & IPW-SY                   & {\cellcolor{black}}\textcolor{white}{0.5752}  & 0.1975                                              & 0.0723                                              & 0.1727                                              & 0.0647                                               & {\cellcolor[rgb]{0.749,0.749,0.749}}\textbf{0.0537} & 0.1775                                               &                                               & IPW-SY                                       & {\cellcolor{black}}\textcolor{white}{0.3070} & {\cellcolor{black}}\textcolor{white}{0.1093}        & {\cellcolor{black}}\textcolor{white}{0.0652}  & 0.0994                                              & 0.0177                                              & 0.0162                                               & {\cellcolor[rgb]{0.749,0.749,0.749}}0.0460           \\
                                 & ModifiedNB               & 0.0816                                        & 0.1944                                              & {\cellcolor[rgb]{0.749,0.749,0.749}}\textbf{0.0113} & {\cellcolor{black}}\textcolor{white}{1.0754}        & 0.2342                                               & 0.2268                                              & {\cellcolor{black}}\textcolor{white}{0.2548}         &                                               & ModifiedNB                                   & 0.0809                                       & {\cellcolor{black}}\textcolor{white}{0.0975}        & 0.0060                                        & {\cellcolor{black}}\textcolor{white}{0.2978}        & 0.0206                                              & 0.0223                                               & {\cellcolor{black}}\textcolor{white}{0.1082}         \\
                                 & 2NB                      & 0.0324                                        & {\cellcolor[rgb]{0.749,0.749,0.749}}\textbf{0.0859} & 0.1303                                              & {\cellcolor{black}}\textcolor{white}{0.2358}        & {\cellcolor{black}}\textcolor{white}{0.3707}         & {\cellcolor{black}}\textcolor{white}{0.3889}        & 0.1604                                               &                                               & 2NB                                          & 0.0054                                       & 0.0565                                              & {\cellcolor{black}}\textcolor{white}{0.0191}  & {\cellcolor{black}}\textcolor{white}{0.1149}        & 0.0103                                              & 0.0096                                               & {\cellcolor[rgb]{0.749,0.749,0.749}}0.0666           \\
                                 & LatentV                  & {\cellcolor{black}}\textcolor{white}{0.0939}  & {\cellcolor{black}}\textcolor{white}{0.2534}        & {\cellcolor[rgb]{0.749,0.749,0.749}}0.0457          & 0.0841                                              & 0.1499                                               & 0.1792                                              & {\cellcolor[rgb]{0.749,0.749,0.749}}\textbf{0.0675}  &                                               & LatentV                                      & {\cellcolor{black}}\textcolor{white}{0.0928} & {\cellcolor{black}}\textcolor{white}{0.1267}        & {\cellcolor{black}}\textcolor{white}{0.0133}  & 0.1082                                              & {\cellcolor{black}}\textcolor{white}{0.0311}        & {\cellcolor{black}}\textcolor{white}{0.0452}         & {\cellcolor[rgb]{0.749,0.749,0.749}}0.0615           \\
                                 & FairMap                  & $\textbackslash$                              & $\textbackslash$                                    & $\textbackslash$                                    & {\cellcolor[rgb]{0.749,0.749,0.749}}\textbf{0.0160} & 0.0662                                               & $\textbackslash$                                    & $\textbackslash$                                     &                                               & FairMap                                      & $\textbackslash$                             & $\textbackslash$                                    & $\textbackslash$                              & {\cellcolor{black}}\textcolor{white}{0.1255}        & {\cellcolor[rgb]{0.749,0.749,0.749}}\textbf{0.0007} & $\textbackslash$                                     & $\textbackslash$                                     \\
                                 & LabelBias-DP             & 0.0816                                        & 0.1672                                              & 0.0617                                              & {\cellcolor{black}}\textcolor{white}{0.6560}        & {\cellcolor{black}}\textcolor{white}{0.4156}         & {\cellcolor[rgb]{0.749,0.749,0.749}}0.1056          & {\cellcolor{black}}\textcolor{white}{1.1349}         &                                               & LabelBias-DP                                 & 0.0809                                       & 0.0387                                              & {\cellcolor{black}}\textcolor{white}{0.0399}  & {\cellcolor[rgb]{0.749,0.749,0.749}}0.0599          & {\cellcolor{black}}\textcolor{white}{0.0545}        & {\cellcolor{black}}\textcolor{white}{0.1654}         & {\cellcolor{black}}\textcolor{white}{0.4018}         \\
                                 & LabelBias-FPR            & 0.0688                                        & 0.1796                                              & {\cellcolor[rgb]{0.749,0.749,0.749}}0.0539          & 0.0925                                              & 0.0972                                               & {\cellcolor[rgb]{0.749,0.749,0.749}}0.1374          & {\cellcolor[rgb]{0.749,0.749,0.749}}0.0985           &                                               & LabelBias-FPR                                & 0.0685                                       & {\cellcolor{black}}\textcolor{white}{0.0965}        & {\cellcolor{black}}\textcolor{white}{0.0151}  & 0.0994                                              & {\cellcolor{black}}\textcolor{white}{0.0364}        & {\cellcolor{black}}\textcolor{white}{0.0569}         & {\cellcolor[rgb]{0.749,0.749,0.749}}0.0421           \\
                                 & LabelBias-EOdds          & 0.0452                                        & {\cellcolor{black}}\textcolor{white}{0.3102}        & {\cellcolor{black}}\textcolor{white}{1.6794}        & 0.1409                                              & 0.0972                                               & {\cellcolor{black}}\textcolor{white}{0.3634}        & 0.1257                                               &                                               & LabelBias-EOdds                              & 0.0423                                       & 0.0169                                              & {\cellcolor{black}}\textcolor{white}{0.7785}  & {\cellcolor[rgb]{0.749,0.749,0.749}}\textbf{0.0244} & {\cellcolor{black}}\textcolor{white}{0.0364}        & {\cellcolor{black}}\textcolor{white}{0.0540}         & {\cellcolor[rgb]{0.749,0.749,0.749}}0.0706           \\
                                 & ARL                      & 0.0762                                        & 0.1825                                              & 0.0609                                              & 0.1650                                              & {\cellcolor{black}}\textcolor{white}{0.2486}         & 0.2888                                              & 0.1735                                               &                                               & ARL                                          & 0.0757                                       & 0.0757                                              & {\cellcolor{black}}\textcolor{white}{0.0127}  & 0.1059                                              & {\cellcolor{black}}\textcolor{white}{0.0314}        & {\cellcolor{black}}\textcolor{white}{0.0300}         & 0.0907                                               \\ 
\cline{2-9}\cline{11-18}
                                 & IFFair-Uniform           & 0.0063                                        & 0.1299                                              & 0.0562                                              & 0.0739                                              & \textbf{0.0209}                                      & 0.2388                                              & 0.2150                                               &                                               & IFFair-Uniform                               & 0.0077                                       & 0.0387                                              & 0.0061                                        & 0.0984                                              & 0.0011                                              & \textbf{0.0041}                                      & 0.1065                                               \\
                                 & IFFair-Diverse           & \textbf{0.0002}                               & 0.1092                                              & 0.0717                                              & 0.0606                                              & 0.0743                                               & 0.1689                                              & 0.1073                                               &                                               & IFFair-Diverse                               & \textbf{0.0014}                              & \textbf{0.0095}                                     & \textbf{0.0057}                               & 0.1018                                              & 0.0059                                              & 0.0054                                               & 0.0742                                               \\
\hline
\end{tabular}
\end{adjustbox}
\end{table}

\textbf{For RQ1.1}, we analyze 28 metric-dataset scenarios. For IFFair-Uniform, the proportion of scenarios where the fairness improvement effect reaches more than 50\% is 17 out of 28 (60.7\%), among which 9 scenarios have an improvement close to 90\%. For IFFair-Diverse, 19 out of 28 scenarios (67.9\%) achieve a fairness improvement of more than 50\%, with 8 scenarios showing an improvement close to 90\%. On average, the average improvement rates of both IFFair-Uniform and IFFair-Diverse across various fairness metrics are $\approx$50\%. In particular, due to its personalized weight adjustment strategy, IFFair-Diverse achieves a better improvement >50\% on LR under all metrics (the average improvement of $\Delta DP$ reaches 53.8\%, $\Delta FPR$ reaches 81.8\%, $\Delta EOdds$ reaches 61.23\%, and $\Delta Err$ reaches 61.37\%). It indicates that \textbf{IFFair can achieve a excellent mitigation in fairness disparity}, obtain fairer decisions just by reweighting samples, without modifying the original data information or the model structure. \textbf{For RQ1.2}, no metrics values on 7 real-world datasets of IFFair are larger than them of vanilla model. It indicateds that \textbf{our methods optimize multiple group fairness metrics without potential conflicts} concerned in \cite{berk2021fairness,du2020fairness}.
\begin{table}[!]
\centering
\caption{Utility comparison of IFFair and baselines on LR. Dark gray cells indicate methods achieve utility improvement and underlines indicate the worst utility.}
\label{tab:util_U&R_LR}
\resizebox{0.75\textwidth}{0.31\textheight}{
\begin{tabular}{|c|c|ccccccc|} 
\hline
\multirow{2}{*}{Utility} & \multirow{2}{*}{Methods} & \multicolumn{7}{c|}{Datasets}                                                                                                                                                                                                                                                                                             \\ 
\cline{3-9}
                         &                          & Bank                                       & German                                     & LSAC                                       & Adult                                      & COMPAS                                     & Comm                                       & MEPS                                        \\ 
\hline
\multirow{15}{*}{$Acc$}  & Original                 & 0.7024                                     & 0.6600                                     & 0.7100                                     & 0.8258                                     & 0.6521                                     & 0.8221                                     & 0.7887                                      \\
                         & Supreesion               & {\cellcolor[rgb]{0.498,0.498,0.498}}0.7046 & {\cellcolor[rgb]{0.498,0.498,0.498}}0.6650 & {\cellcolor[rgb]{0.498,0.498,0.498}}0.7129 & 0.8254                                     & {\cellcolor[rgb]{0.498,0.498,0.498}}0.6529 & 0.8170                                     & {\cellcolor[rgb]{0.498,0.498,0.498}}0.7912  \\
                         & IPW-S                    & \uline{0.5354}                             & {\cellcolor[rgb]{0.498,0.498,0.498}}0.6650 & 0.6974                                     & 0.8225                                     & {\cellcolor[rgb]{0.498,0.498,0.498}}0.6577 & 0.8195                                     & 0.7577                                      \\
                         & IPW-SY                   & 0.6250                                     & {\cellcolor[rgb]{0.498,0.498,0.498}}0.6700 & 0.6538                                     & 0.7773                                     & {\cellcolor[rgb]{0.498,0.498,0.498}}0.6569 & 0.7970                                     & 0.7097                                      \\
                         & ModifiedNB               & {\cellcolor[rgb]{0.498,0.498,0.498}}0.7024 & {\cellcolor[rgb]{0.498,0.498,0.498}}0.7000 & 0.6565                                     & \uline{0.7061}                             & 0.6464                                     & 0.7669                                     & \uline{0.4975}                              \\
                         & 2NB                      & {\cellcolor[rgb]{0.498,0.498,0.498}}0.9679 & {\cellcolor[rgb]{0.498,0.498,0.498}}0.7250 & {\cellcolor[rgb]{0.498,0.498,0.498}}0.9114 & {\cellcolor[rgb]{0.498,0.498,0.498}}0.8377 & 0.6480                                     & 0.8095                                     & {\cellcolor[rgb]{0.498,0.498,0.498}}0.8566  \\
                         & LatentV                  & {\cellcolor[rgb]{0.498,0.498,0.498}}0.7124 & {\cellcolor[rgb]{0.498,0.498,0.498}}0.7050 & {\cellcolor[rgb]{0.498,0.498,0.498}}0.8866 & {\cellcolor[rgb]{0.498,0.498,0.498}}0.8396 & \uline{0.6310}                             & 0.8170                                     & {\cellcolor[rgb]{0.498,0.498,0.498}}0.8342  \\
                         & CostFree-DP              & \textbackslash{}                           & 0.6450                                     & \textbackslash{}                           & 0.8074                                     & 0.6496                                     & 0.7970                                     & \textbackslash{}                            \\
                         & CostFree-FPR             & \textbackslash{}                           & {\cellcolor[rgb]{0.498,0.498,0.498}}0.6600 & \textbackslash{}                           & {\cellcolor[rgb]{0.498,0.498,0.498}}0.8260 & 0.6496                                     & 0.8195                                     & \textbackslash{}                            \\
                         & LabelBias-DP             & {\cellcolor[rgb]{0.498,0.498,0.498}}0.7024 & {\cellcolor[rgb]{0.498,0.498,0.498}}0.6650 & {\cellcolor[rgb]{0.498,0.498,0.498}}0.7132 & 0.7656                                     & 0.6318                                     & \uline{0.7293}                             & 0.5578                                      \\
                         & LabelBias-FPR            & {\cellcolor[rgb]{0.498,0.498,0.498}}0.7058 & 0.6500                                     & {\cellcolor[rgb]{0.498,0.498,0.498}}0.7124 & 0.8243                                     & {\cellcolor[rgb]{0.498,0.498,0.498}}0.6521 & 0.7870                                     & 0.7808                                      \\
                         & LabelBias-EOdds          & {\cellcolor[rgb]{0.498,0.498,0.498}}0.7024 & \uline{0.6150}                             & \uline{0.5738}                             & 0.7336                                     & {\cellcolor[rgb]{0.498,0.498,0.498}}0.6521 & 0.7970                                     & 0.7056                                      \\
                         & ARL                      & 0.6980                                     & 0.6500                                     & 0.7095                                     & {\cellcolor[rgb]{0.498,0.498,0.498}}0.8260 & {\cellcolor[rgb]{0.498,0.498,0.498}}0.6537 & 0.8195                                     & 0.7865                                      \\ 
\hhline{|~--------|}
                         & IFFair-Uniform           & {\cellcolor[rgb]{0.498,0.498,0.498}}0.7168 & {\cellcolor[rgb]{0.498,0.498,0.498}}0.6650 & {\cellcolor[rgb]{0.498,0.498,0.498}}0.7113 & 0.8254                                     & 0.6342                                     & 0.8020                                     & {\cellcolor[rgb]{0.498,0.498,0.498}}0.7887  \\
                         & IFFair-Diverse           & {\cellcolor[rgb]{0.498,0.498,0.498}}0.7046 & {\cellcolor[rgb]{0.498,0.498,0.498}}0.6600 & {\cellcolor[rgb]{0.498,0.498,0.498}}0.7782 & {\cellcolor[rgb]{0.498,0.498,0.498}}0.8309 & 0.6440                                     & 0.8120                                     & {\cellcolor[rgb]{0.498,0.498,0.498}}0.8117  \\ 
\hline
\multirow{13}{*}{$F1$}   & Original                 & 0.4979                                     & 0.6486                                     & 0.5894                                     & 0.7729                                     & 0.6491                                     & 0.8218                                     & 0.7055                                      \\
                         & Supreesion               & {\cellcolor[rgb]{0.498,0.498,0.498}}0.4992 & {\cellcolor[rgb]{0.498,0.498,0.498}}0.6556 & {\cellcolor[rgb]{0.498,0.498,0.498}}0.5924 & 0.7718                                     & {\cellcolor[rgb]{0.498,0.498,0.498}}0.6499 & 0.8166                                     & {\cellcolor[rgb]{0.498,0.498,0.498}}0.7084  \\
                         & IPW-S                    & \uline{0.3929}                             & {\cellcolor[rgb]{0.498,0.498,0.498}}0.6531 & 0.5812                                     & {\cellcolor[rgb]{0.498,0.498,0.498}}0.7731 & {\cellcolor[rgb]{0.498,0.498,0.498}}0.6540 & 0.8191                                     & 0.6778                                      \\
                         & IPW-SY                   & 0.4509                                     & {\cellcolor[rgb]{0.498,0.498,0.498}}0.6589 & 0.5532                                     & 0.7455                                     & {\cellcolor[rgb]{0.498,0.498,0.498}}0.6548 & 0.7967                                     & 0.6427                                      \\
                         & ModifiedNB               & {\cellcolor[rgb]{0.498,0.498,0.498}}0.4979 & {\cellcolor[rgb]{0.498,0.498,0.498}}0.6875 & 0.5543                                     & \uline{0.6807}                             & 0.6442                                     & 0.7669                                     & \uline{0.4803}                              \\
                         & 2NB                      & {\cellcolor[rgb]{0.498,0.498,0.498}}0.6696 & {\cellcolor[rgb]{0.498,0.498,0.498}}0.6925 & {\cellcolor[rgb]{0.498,0.498,0.498}}0.6524 & 0.7653                                     & 0.6403                                     & 0.8093                                     & {\cellcolor[rgb]{0.498,0.498,0.498}}0.7139  \\
                         & LatentV                  & {\cellcolor[rgb]{0.498,0.498,0.498}}0.5039 & {\cellcolor[rgb]{0.498,0.498,0.498}}0.6907 & {\cellcolor[rgb]{0.498,0.498,0.498}}0.6641 & {\cellcolor[rgb]{0.498,0.498,0.498}}0.7750 & \uline{0.6270}                             & 0.8165                                     & {\cellcolor[rgb]{0.498,0.498,0.498}}0.7257  \\
                         & LabelBias-DP             & {\cellcolor[rgb]{0.498,0.498,0.498}}0.4979 & {\cellcolor[rgb]{0.498,0.498,0.498}}0.6568 & {\cellcolor[rgb]{0.498,0.498,0.498}}0.5922 & 0.6926                                     & 0.6275                                     & \uline{0.7205}                             & 0.5109                                      \\
                         & LabelBias-FPR            & {\cellcolor[rgb]{0.498,0.498,0.498}}0.4999 & 0.6383                                     & {\cellcolor[rgb]{0.498,0.498,0.498}}0.5916 & 0.7710                                     & 0.6482                                     & 0.7848                                     & 0.7007                                      \\
                         & LabelBias-EOdds          & {\cellcolor[rgb]{0.498,0.498,0.498}}0.4979 & \uline{0.6042}                             & \uline{0.4645}                             & 0.7060                                     & 0.6482                                     & 0.7969                                     & 0.6444                                      \\
                         & ARL                      & 0.4952                                     & 0.6408                                     & 0.5893                                     & 0.7718                                     & {\cellcolor[rgb]{0.498,0.498,0.498}}0.6498 & 0.8191                                     & 0.7042                                      \\ 
\hhline{|~--------|}
                         & IFFair-Uniform           & {\cellcolor[rgb]{0.498,0.498,0.498}}0.5067 & {\cellcolor[rgb]{0.498,0.498,0.498}}0.6544 & {\cellcolor[rgb]{0.498,0.498,0.498}}0.5911 & 0.7684                                     & 0.6316                                     & 0.8017                                     & {\cellcolor[rgb]{0.498,0.498,0.498}}0.7055  \\
                         & IFFair-Diverse           & {\cellcolor[rgb]{0.498,0.498,0.498}}0.4992 & {\cellcolor[rgb]{0.498,0.498,0.498}}0.6499 & {\cellcolor[rgb]{0.498,0.498,0.498}}0.6277 & 0.7675                                     & 0.6402                                     & 0.8119                                     & {\cellcolor[rgb]{0.498,0.498,0.498}}0.7179  \\ 
\hline
\multirow{14}{*}{$AUC$}  & Original                 & 0.9738                                     & 0.8047                                     & 0.8663                                     & 0.8870                                     & 0.7063                                     & 0.8972                                     & 0.8375                                      \\
                         & Supreesion               & 0.9735                                     & {\cellcolor[rgb]{0.498,0.498,0.498}}0.8047 & {\cellcolor[rgb]{0.498,0.498,0.498}}0.8665 & 0.8869                                     & {\cellcolor[rgb]{0.498,0.498,0.498}}0.7064 & {\cellcolor[rgb]{0.498,0.498,0.498}}0.8986 & 0.8360                                      \\
                         & IPW-S                    & 0.7643                                     & 0.7968                                     & 0.8523                                     & 0.8860                                     & {\cellcolor[rgb]{0.498,0.498,0.498}}0.7080 & {\cellcolor[rgb]{0.498,0.498,0.498}}0.8976 & 0.8009                                      \\
                         & IPW-SY                   & 0.9347                                     & 0.8001                                     & {\cellcolor[rgb]{0.498,0.498,0.498}}0.8726 & {\cellcolor[rgb]{0.498,0.498,0.498}}0.8938 & 0.6988                                     & 0.8828                                     & 0.8197                                      \\
                         & ModifiedNB               & {\cellcolor[rgb]{0.498,0.498,0.498}}0.9738 & 0.8016                                     & 0.8626                                     & 0.8194                                     & 0.6937                                     & 0.8661                                     & 0.8066                                      \\
                         & 2NB                      & 0.8832                                     & 0.7621                                     & {\cellcolor[rgb]{0.498,0.498,0.498}}0.8739 & {\cellcolor[rgb]{0.498,0.498,0.498}}0.8888 & 0.6973                                     & 0.8845                                     & {\cellcolor[rgb]{0.498,0.498,0.498}}0.8384  \\
                         & LatentV                  & 0.9682                                     & 0.7902                                     & 0.7034                                     & 0.8782                                     & 0.6882                                     & 0.8929                                     & 0.7946                                      \\
                         & FairMap                  & \uline{0.5080}                             & \uline{0.4236}                             & \uline{0.5170}                             & \uline{0.5389}                             & \uline{0.5156}                             & \uline{0.4241}                             & \uline{0.5069}                              \\
                         & LabelBias-DP             & {\cellcolor[rgb]{0.498,0.498,0.498}}0.9738 & 0.7771                                     & 0.8661                                     & 0.8027                                     & 0.6813                                     & 0.8060                                     & 0.6911                                      \\
                         & LabelBias-FPR            & 0.9734                                     & 0.7953                                     & {\cellcolor[rgb]{0.498,0.498,0.498}}0.8665 & 0.8862                                     & 0.6995                                     & 0.8744                                     & 0.8231                                      \\
                         & LabelBias-EOdds          & 0.9679                                     & 0.7021                                     & {\cellcolor[rgb]{0.498,0.498,0.498}}0.7191 & 0.8712                                     & 0.6995                                     & 0.8689                                     & 0.8306                                      \\
                         & ARL                      & 0.9735                                     & 0.8022                                     & 0.8662                                     & 0.8865                                     & {\cellcolor[rgb]{0.498,0.498,0.498}}0.7066 & 0.8969                                     & 0.8352                                      \\ 
\hhline{|~--------|}
                         & IFFair-Uniform           & 0.9702                                     & 0.7783                                     & {\cellcolor[rgb]{0.498,0.498,0.498}}0.8665 & 0.8843                                     & 0.6889                                     & 0.8768                                     & {\cellcolor[rgb]{0.498,0.498,0.498}}0.8375  \\
                         & IFFair-Diverse           & 0.9655                                     & 0.7824                                     & 0.8250                                     & 0.8796                                     & 0.6884                                     & 0.8728                                     & 0.8196                                      \\
\hline
\end{tabular}}
\end{table}

\subsubsection{RQ2: How well does IFFair perform compared with baselines?} In this question, we compare IFFair performance with baselines to observe whether our method can perform better than these algorithms, which is helpful to provide insights in selecting methods with different demands. RQ2 is going to be discussed from two aspects: How well does IFFair perform compared with baselines on fairness (\textbf{RQ2.1}) and utility (\textbf{RQ2.2})?

\textbf{For RQ2.1}, we analyze comparisons in Table \ref{tab:fair_U&D_LR} between IFFair and baselines on 4 fairness metrics: (1) IFFair outperforms baselines (254/304=83.5\%) in most cases. Specifically, for 13 implementations of baselines involved in the comparison on $\Delta DP$ and $\Delta FPR$, 11 on $\Delta DP$ and $\Delta FPR$ respectively, there are 304 valid results, where only 50 (light gray cells) perform a litte better than ours. (2) Only IFFair doesn't exacerbate bias (black cells) of $Original$, that is not achieved by other baselines. (3) Across all datasets and fairness metrics, the proportion of our method obtaining the optimal solution for fairness optimization is 10/28=35.7\%, which is the highest among all methods.

\textbf{For RQ2.2}, we analyze results between IFFair and baselines on 3 utility metrics in Table \ref{tab:util_U&R_LR}. To show the impact on performance loss, we compare both IFFair and baselines with the vanilla classifier without any debiased strategy. The comparisons show: (1) Most baselines cause a decline in classification performance while improving fairness. When analysing dataset-metrics cells, the decline proportions (black cells) in $Acc$, $F1$, $AUC$ are 50/92 = 54.3\%, 46/84 = 54.8\%, 73/91 = 80.2\% respectively. When analysing by a specific method, there is no method that can achieve utility improvement across all datasets and metrics. (2) In 21 metrics-datasets cases, IFFair never seriously damages the performance of $Original$ (without underlines). (3) For utility improvemnet (dark gray), the propotion of IFFair is 19/42=45.2\% that is highr than 5 out of the 6 baselines. It means that IFFair can always maintain even improve original utility.

\textbf{The conclusion of RQ2 is:} IFFair significantly improves machine learning fairness and outperforms baseline methods in most cases. It doesn't servely affect the classification performance of the original model, and its performance loss is smaller than most baselines even decreases in most cases.

\subsubsection{RQ3: How general is IFFair performs on different basic networks?}
To verify the generality of IFFair, we also conduct experiments on DNN. It employs a two-layer fully connected multi-layer perceptron as the feature extractor, adopts the ReLU activation function \cite{nair2010rectified} to achieve non-linear transformations, and uses a logistic regression layer as the classifier. The inverse operation of the Hessian matrix requires the loss function to be convex. Therefore, when calculating the influence function of the DNN, we approximate the overall influence by computing the influence of the last layer without a non-linear activation function.
\begin{table}[h]
\centering
\caption{Comparisons of fairness and utility between IFFair and baselines on DNN.}\label{tab:IFFair_DNN}
\resizebox{0.7\textwidth}{0.28\textheight}{
\begin{tabular}{|c|c|ccccccc|} 
\hline
\multirow{2}{*}{Metrics} & \multirow{2}{*}{Methods} & \multicolumn{7}{c|}{Datasets}                                                                                                                             \\ 
\cline{3-9}
                         &                          & Bank                                       & German          & LSAC            & Adult            & COMPAS           & Comm            & MEPS             \\ 
\hline
\multirow{7}{*}{$\Delta DP$}      & Original                 & 0.0066 & 0.1346          & 0.0386          & 0.1925           & 0.1598           & 0.4319          & 0.1160           \\
                         & Supreesion               & 0.0063                                     & 0.1687          & 0.0389          & 0.1898           & 0.1092           & 0.3982          & 0.1053           \\
                         & IPW-S                    & {\cellcolor[rgb]{0.749,0.749,0.749}}\textbf{0.0053}& 0.1742          & 0.0509          & 0.1853           & 0.1183           & 0.4078          & 0.1235           \\
                         & IPW-SY                   & 0.0125                                     & 0.2532          & 0.0459          & 0.2016           & {\cellcolor[rgb]{0.749,0.749,0.749}}\textbf{0.0373}& {\cellcolor[rgb]{0.749,0.749,0.749}}\textbf{0.3197}& 0.0870           \\
                         & CostFree-DP              & \textbackslash                                       & \textbackslash            & \textbackslash            & {\cellcolor[rgb]{0.749,0.749,0.749}}0.1682& 0.1030           & \textbackslash            & \textbackslash             \\
                         & CostFree-FPR             & \textbackslash                                       & \textbackslash            & \textbackslash            & {\cellcolor[rgb]{0.749,0.749,0.749}}\textbf{0.1596}& 0.1106           & \textbackslash            & \textbackslash             \\ 
\cline{2-9}
                         & IFFair-Diverse           & -0.0062& \textbf{0.0397} & \textbf{0.0286} & 0.1729           & 0.0839           & 0.3823          & \textbf{0.0713}  \\ 
\hline
\multirow{7}{*}{$\Delta FPR$}     & Original                 & 0.1200                                     & 0.1667          & 0.0216          & 0.1133           & 0.0800           & 0.1986          & 0.1526           \\
                         & Supreesion               & 0.0367                                     & 0.1217          & 0.0142          & 0.0966           & 0.0347           & 0.1730          & 0.1563           \\
                         & IPW-S                    & 0.0533                                     & 0.2459          & 0.0263          & 0.1088           & 0.0590           & 0.2063          & 0.1926           \\
                         & IPW-SY                   & -0.0367                                    & 0.2887          & 0.0202          & {\cellcolor[rgb]{0.749,0.749,0.749}}\textbf{-0.0039}& {\cellcolor[rgb]{0.749,0.749,0.749}}\textbf{-0.0169}& {\cellcolor[rgb]{0.749,0.749,0.749}}\textbf{0.1304}& -0.0622          \\
                         & CostFree-DP              & \textbackslash                                       & \textbackslash            & \textbackslash            & {\cellcolor[rgb]{0.749,0.749,0.749}}0.0564& 0.0527           & \textbackslash            & \textbackslash             \\
                         & CostFree-FPR             & \textbackslash                                       & \textbackslash            & \textbackslash            & {\cellcolor[rgb]{0.749,0.749,0.749}}0.0211& 0.0436           & \textbackslash            & \textbackslash             \\ 
\cline{2-9}
                         & IFFair-Diverse           & \textbf{-0.0100}                           & \textbf{0.0230} & \textbf{0.0138} & 0.0670           & 0.0266           & 0.1381          & \textbf{0.0460}  \\ 
\hline
\multirow{5}{*}{$\Delta EOdds$}   & Original                 & 0.1227                                     & 0.1960          & 0.1598          & 0.2019           & 0.2819           & 0.3690          & 0.2033           \\
                         & Supreesion               & 0.0388                                     & 0.3358          & 0.2465          & 0.1800           & 0.1838           & 0.3308          & 0.1884           \\
                         & IPW-S                    & 0.0572                                     & 0.3615          & 0.2626          & 0.1942           & 0.1946           & 0.3723          & 0.2459           \\
                         & IPW-SY                   & 0.0502                                     & 0.4284          & 0.1332          & {\cellcolor[rgb]{0.749,0.749,0.749}}\textbf{0.1089}& {\cellcolor[rgb]{0.749,0.749,0.749}}\textbf{0.0755}& {\cellcolor[rgb]{0.749,0.749,0.749}}\textbf{0.2236}& 0.0983           \\ 
\cline{2-9}
                         & IFFair-Diverse           & \textbf{0.0174}                            & \textbf{0.0523} & \textbf{0.1241} & 0.1420           & 0.1177           & 0.2770          & \textbf{0.0665}  \\ 
\hline
\multirow{5}{*}{$\Delta Err$}     & Original                 & 0.0040                                     & 0.1361          & 0.0232          & 0.1148           & 0.0192           & 0.0406          & 0.0747           \\
                         & Supreesion               & 0.0104                                     & 0.0442          & {\cellcolor[rgb]{0.749,0.749,0.749}}0.0131& {\cellcolor[rgb]{0.749,0.749,0.749}}0.1130& 0.0280           & \textbf{0.0214} & {\cellcolor[rgb]{0.749,0.749,0.749}}\textbf{0.0623}\\
                         & IPW-S                    & 0.0176                                     & 0.2007          & 0.0209          & {\cellcolor[rgb]{0.749,0.749,0.749}}0.1125& {\cellcolor[rgb]{0.749,0.749,0.749}}\textbf{0.0032}& 0.0357          & 0.0731           \\
                         & IPW-SY                   & 0.0661                                     & 0.1650          & {\cellcolor[rgb]{0.749,0.749,0.749}}\textbf{0.0067}& {\cellcolor[rgb]{0.749,0.749,0.749}}\textbf{0.0791}& 0.0592           & 0.0388          & {\cellcolor[rgb]{0.749,0.749,0.749}}0.0428\\ 
\cline{2-9}
                         & IFFair-Diverse           & \textbf{0.0029}                            & \textbf{0.0413} & 0.0193          & 0.1134           & 0.0106           & \textbf{0.0214} & 0.0710           \\ 
\hline
\multirow{7}{*}{$Acc$}     & Original                 & 0.9690                                     & \underline{0.7600}  & 0.9064          & 0.8424           & 0.6204           & 0.8095          & 0.8522           \\
                         & Supreesion               & 0.9599                                     & 0.7697          & 0.9059          & 0.8428           & 0.6127           & 0.7875          & 0.8516           \\
                         & IPW-S                    & 0.9613                                     & 0.8020          & 0.9056          & 0.8439           & 0.6081           & 0.7827          & 0.8459           \\
                         & IPW-SY                   & \underline{0.9270}                             & \underline{0.7600}  & \underline{0.7705}  & \underline{0.8026}   & \underline{0.5919}   & \underline{0.7438}  & \underline{0.7852}   \\
                         & CostFree-DP              & \textbackslash                                       & \textbackslash            & \textbackslash            & 0.8422           & 0.6103           & \textbackslash            & \textbackslash             \\
                         & CostFree-FPR             & \textbackslash                                       & \textbackslash            & \textbackslash            & 0.8402           & 0.6118           & \textbackslash            & \textbackslash             \\ 
\cline{2-9}
                         & IFFair-Diverse           & 0.9679                                     & 0.7900          & 0.9066          & 0.8414           & 0.6172           & 0.7945          & 0.8522           \\ 
\hline
\multirow{5}{*}{$F1$}      & Original                 & 0.5803                                     & \underline{0.7195}  & 0.6471          & 0.7775           & 0.6162           & 0.8093          & 0.7112           \\
                         & Supreesion               & 0.8160                                     & 0.7352          & 0.6375          & 0.7775           & 0.6059           & 0.7874          & 0.7128           \\
                         & IPW-S                    & 0.7620                                     & 0.7753          & 0.6447          & 0.7789           & 0.6047           & 0.7825          & 0.7009           \\
                         & IPW-SY                   & \underline{0.5577}                             & 0.7366          & \underline{0.6301}  & \underline{0.7407}   & \underline{0.5910}   & \underline{0.7437}  & \underline{0.6968}   \\ 
\cline{2-9}
                         & IFFair-Diverse           & 0.6460                                     & 0.7476          & 0.6413          & 0.7743           & 0.6142           & 0.7944          & 0.7065           \\ 
\hline
\multirow{5}{*}{$AUC$}     & Original                 & 0.9431                                     & 0.7996          & 0.8690          & 0.8957           & 0.6564           & 0.8829          & 0.8338           \\
                         & Supreesion               & \underline{0.7036}                             & 0.8420          & 0.8786          & 0.8965           & 0.6511           & 0.8757          & 0.8305           \\
                         & IPW-S                    & 0.8877                                     & 0.8407          & 0.8712          & 0.8959           & 0.6380           & 0.8820          & 0.8275           \\
                         & IPW-SY                   & 0.9035                                     & 0.8373          & 0.8710          & \underline{0.8908}   & \underline{0.6302}   & \underline{0.8534}  & \underline{0.8142}   \\ 
\cline{2-9}
                         & IFFair-Diverse           & 0.9167                                     & \underline{0.7707}  & \underline{0.8636}  & 0.8931           & 0.6483           & 0.8643          & 0.8158           \\
\hline
\end{tabular}
}
\end{table}

Table \ref{tab:IFFair_DNN} show the fairness and utility performances on DNN that is similar with LR as expected. Based on 106 fairness baseline results, the proportion of cases where IFFair wins reaches 85/106=80.2\%. In addition, in 28 metrics-datasets fairness scenarios, the probability that IFFair obtains the optimal solution is 50\%, which keeps the significant advantage among all methods. IPW-SY is slightly inferior to our fairness optimization, which covers 10 scenarios. Regarding the utility results, the worst performance is recorded with underlines. For the 2nd optimal fairness method IPW-SY, it has the worst performance where 17/22 underlines belongs to IPW-SY. Although IFFair causes a few damage cases to utility, it only accounts for 2/22 and the loss is small (0.6\%-3.6\%).

\textbf{RQ3 concludes that} IFFair is generalizable, which can be applied to LR and DNN with adapting capabilities of linear and non-linear transforming.
\begin{figure}[!]
\includegraphics[width=\textwidth]{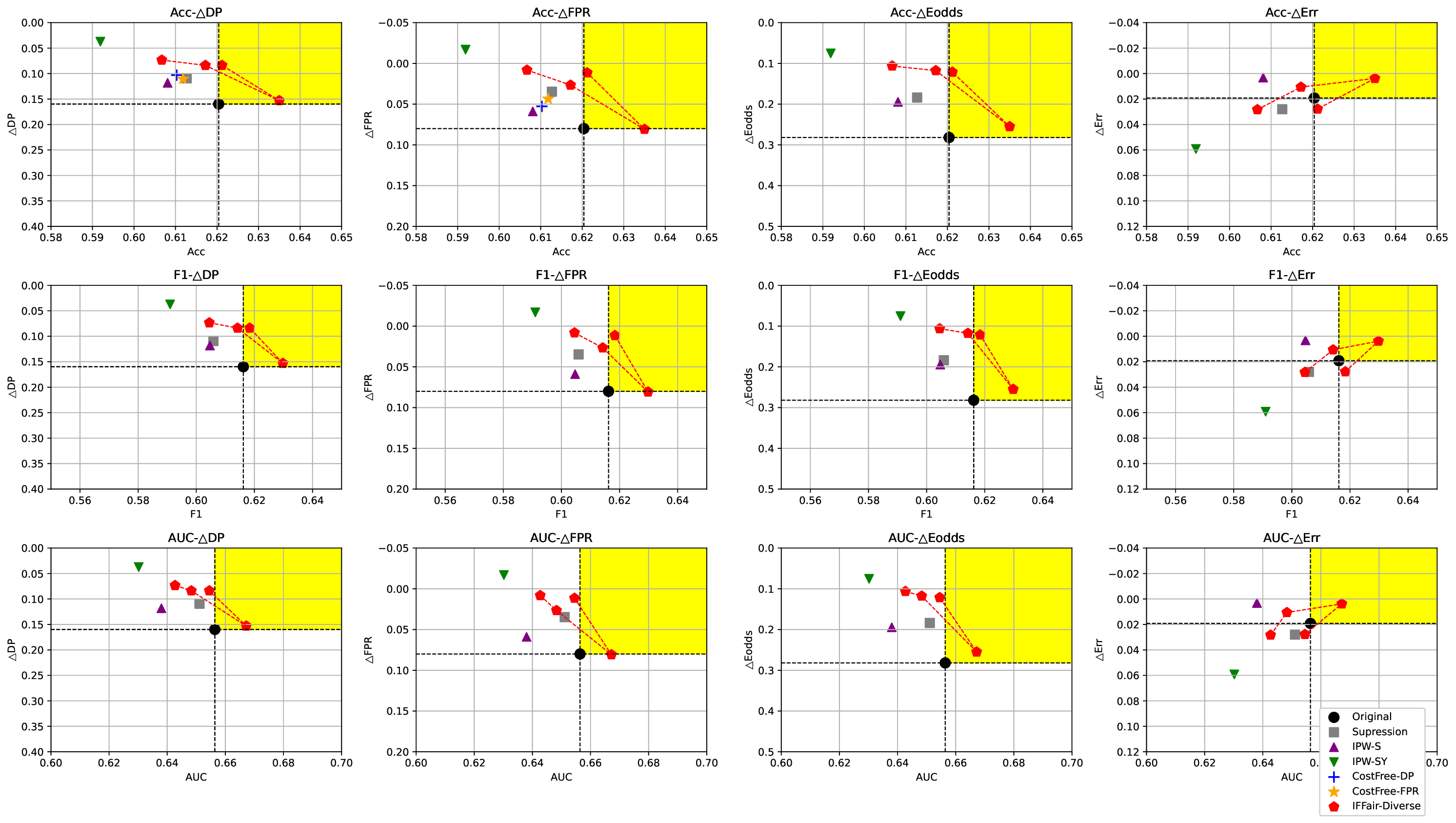}
\caption{Utility-fairness trade-off of IFFair and baselines on DNN and COMPAS dataset.} \label{fig:dnn_compas_trade}
\end{figure}

\subsubsection{RQ4: Trade-off between fairness and utility}
We analyze 12 fairness-utility trade-offs (4 fairness and 3 utility metrics) in this section. Because IFFair-Diverse can be applied on both LR and DNN, and its adaptive reweighting strategy provides better potential for fairness optimization, we evaluate its trade-off as the representation of IFFair. Specifically, our method explicitly imposes fairness and utility constraints as Equation \eqref{eq:IFFair-Diverse_constraint}, where we set two hyperparameters $\lambda_f$ and $\lambda_u$ to adjust fairness and utility levels respectively. Both parameters are initialized to 1 indicating no constraints are set. We first decrease $\lambda_u$ in a certain interval to maintain the primary classification goal. Then we adjust $\lambda_f$ to minimize the loss of bias while preserving utility.

We plot Figure \ref{fig:dnn_compas_trade} to show trade-off performance. Most pre-processing methods without explicit constraints are drawn as a point while IFFair is plotted based on changing of $\lambda_f$ with fixed $\lambda_u$. Take account of the space, we mainly show experiment on the complex model DNN and COMPAS adopted in all baselines. For other conditions the results are similar: (1) It's known that the closer the utility value is to 1, the better performance the model has, and the closer the absolute fairness gap is to 0, the fairer the classification is. Therefore, methods closer to the upper right achieve a better trade-off. Red points are more concentrated in the upper-right corner of each subfigure, so \textbf{the trade-off of IFFair is better} than others. (2) Based on the utility and fairness of the vanilla model, we draw horizontal and vertical lines divide the space into 4 regions. The methods falling into the yellow region indicate dual optimizations of utility and fairness. It's easy to find adjusting $\lambda_u$ and $\lambda_f$ always help \textbf{IFFair achieves dual optimization}. (3) In contrast, other methods perform comparable fairness optimization of IFFair cause worse utility damage. Some even fall into the worse-pairwise region (lower left corner), which means that they damage both fairness and utility.

\section{Related Work}
\subsection{Bias Mitigation}
According to different stages of training, fairness algorithms can be divided into pre-processing, in-processing and post-processing methods. Pre-processing methods detect and mitigate the underlying bias in data before training \cite{pessach2022review}. An intuitive method is to delete sensitive attribute \cite{grgic2016case}. Feldman et al. introduced a feature-adjusting disparate impact remover to equalize marginal distributions across groups \cite{feldman2015certifying}. \cite{zhang2017achieving} extracted causal relationships in training data based on causal graphs and modified their labels. Instead of modifying features, \cite{burnaev2015influence} improved fairness by reweighting data distribution. In-processing methods consider fairness during training by modifying objective functions or imposing constraints \cite{mehrabi2021survey}. \cite{beutel2019putting} adopted absolute correlation to improve equal opportunity. CAF \cite{yang2024making} mitigated fairness disparity in predictions across groups based on correlation alignment. Madras et al. used representation learning to mitigate the downstream unfair predictions \cite{madras2018learning}. CFair mitigated bias based on the balanced error rate and conditional representations \cite{zhao2019conditional}. Post-processing methods improve fairness by modifying the prediction after training \cite{d2017conscientious}. kamiran et al. \cite{kamiran2009classifying} adjusted the leaf labels of the decision tree to obtain an unbiased classifier, while \cite{fish2016confidence} changed the decision boundaries of protected groups.

\subsection{Influence Function}
Influence function is used to approximate the actual impact of removing a sample without retraining models\cite{cook1980characterizations,law1986robust}. Related researches of influence functions in ML are not extensive, mainly focusing on robustness and explainability: For robustness, the influence function is widely used in cross-validation optimization\cite{Alam2010ACS,liu2014efficient,giordano2019swiss}. \cite{Alam2010ACS} proposed robust kernel covariance and cross-covariance operators based on influence function to overcome the sensitivity to dirty data. Liu et al. approximated a cross-validation based on the Bouligand influence function, which only requires the algorithm once \cite{liu2014efficient}. \cite{giordano2019swiss} designed a linear approximation method based on the influence function to reduce the dependence of the fitting process on weights. For explainability, \cite{kong2021understanding} constructed counterfactual questions answered by influence functions, and explored the impact of the training samples they revealed on classical unsupervised learning methods. \cite{chen2022characterizing} used the influence function to explain the decision-making from graph CNNs.

\section{Conclusion}
We propose a reweighting method IFFair based on influence function, which improves fairness by quantifying and reducing the influence disparity between groups. Besides, we conduct trade-off constraints on IFFair to balance the relationship between fairness and utility. Experiment on 2 networks, 7 datasets, 4 fairness metrics and 3 utility metrics show that compared with previous pre-processing methods, IFFair not only achieves non-conflicting fairness optimization on fairness-datasets cross pairs, but also achieves a better trade-off in 12 utility-fairness cross scenarios.

\begin{credits}
\subsubsection{\discintname}
The authors have no competing interests to declare that are relevant to the content of this article.
\end{credits}

%
%
%
\bibliographystyle{splncs04}
\bibliography{mybib}

\end{document}